\documentclass[runningheads]{llncs}

\usepackage{eccv}

\usepackage{eccvabbrv}

\usepackage{graphicx}
\usepackage{booktabs}

\usepackage[accsupp]{axessibility}  

\usepackage{multirow}
\usepackage[normalem]{ulem}
\usepackage{graphicx}
\usepackage{makecell}

\usepackage{multirow}
\usepackage{pifont}
\usepackage{wrapfig}

\renewcommand{\thedefinition}{\arabic{definition}}
\newcommand{\mydefinition}[1]{%
  \refstepcounter{definition}
  \textbf{Definition \thedefinition}\label{#1}%
}

\usepackage{hyperref}

\usepackage{orcidlink}

\begin{document}


\title{Medical AI Encodes a ``Feeling of Error'': Verifying Cancer Segmentation via Internal Concepts}




\titlerunning{Verifying Cancer Segmentation via Internal Concepts}

\author{Mengmeng Ma\inst{1}\orcidlink{0000-0002-2804-2718} \and
Yunxiang Peng\inst{2}\orcidlink{0009-0000-1824-970X} \and
Tang Li\inst{2}\orcidlink{0000-0002-3134-4151} \and Lu Lin\inst{3} \and Binsheng Zhao\inst{3} \and Oguz Akin\inst{3} \and Xi Peng\inst{1}\orcidlink{0000-0002-7772-001X}}

\authorrunning{M.~Ma et al.}

\institute{University of Virginia, Charlottesville, VA, USA \and
University of Delaware, Newark, DE, USA \and
 Memorial Sloan Kettering Cancer Center, New York City, NY, USA\\
\email{\{wkb9cb,naq5rd\}@virginia.edu}}

\maketitle

\begin{abstract}
Cancer segmentation models can fail silently, generating plausible but incorrect masks that risk missed findings or unnecessary biopsies. A critical question arises: Do AI models ``know'' when they are wrong, and if so, can we use the signal to predict their own failures? Humans do have a ``Feeling of Error'' (FOE): a spontaneous sense of unease that flags a potential error during thinking. We investigate whether cancer segmentation models exhibit an analogous internal signal. Unlike output-level cues (e.g., prediction confidence or uncertainty), which offer no insight into why a failure occurs and suffer from a sensitivity–quality tradeoff where high detection sensitivity could degrade overall segmentation quality. We instead propose to capture the model's FOE from its inner workings. Using mechanistic interpretability tools,  specifically Sparse Autoencoders, we decompose internal neural activations into a dictionary of human-interpretable concepts and show that failure cases exhibit a distinct latent signature: fewer active concepts with lower activation magnitudes compared to successful segmentation. By training a classifier on these concept activations, we achieve accurate failure detection along with explanations for the model's mistakes. Experiments on prostate, pancreatic, and brain cancer segmentation demonstrate that our approach outperforms output-based methods in failure detection while preserving segmentation quality. Code is abailable at \href{https://github.com/deep-real/CancerSegFailure}{https://github.com/deep-real/CancerSegFailure}.

\end{abstract}    
\section{Introduction}
\label{sec:intro}

The promise of AI in clinical oncology is often undermined by \textit{silent failure}. Even high-performing models can generate anatomically plausible yet incorrect masks that risk missed findings or unnecessary biopsies~\cite{gonzalez2022distance,hosny2022clinical}.
Although clinicians remain the last line of defense, manually verifying numerous AI-generated masks under heavy workloads imposes a significant cognitive burden~\cite{liu2024burnout}. This raises a critical question: \textit{Do AI models ``know'' when they are wrong?} If so, can we use this signal to predict the silent failures? Interestingly, humans do, through what psychologist called ``Feeling of Error'' (FOE)~\cite{gangemi2015foe,wessel2012error}: a spontaneous sensation of cognitive uneasiness arising from conflict detection during thinking, effectively acting as an internal signal to flag potential error. We ask whether state-of-the-art ViT-based segmentation models~\cite{ma2024medsam,zhao2025foundation,cheng2023sam} exhibit an analogous FOE-like signal. Identifying such signal would turn segmentation from a black-box output into an auditable decision process. By tracing and examining the corresponding FOE signals, we can automatically flag high-risk cases for inspection or provide the justifications clinicians need to trust, refine, or override AI predictions.

\begin{wrapfigure}{r}{0.52\linewidth}
  \vspace{-20pt}   
  \centering
  \includegraphics[width=\linewidth]{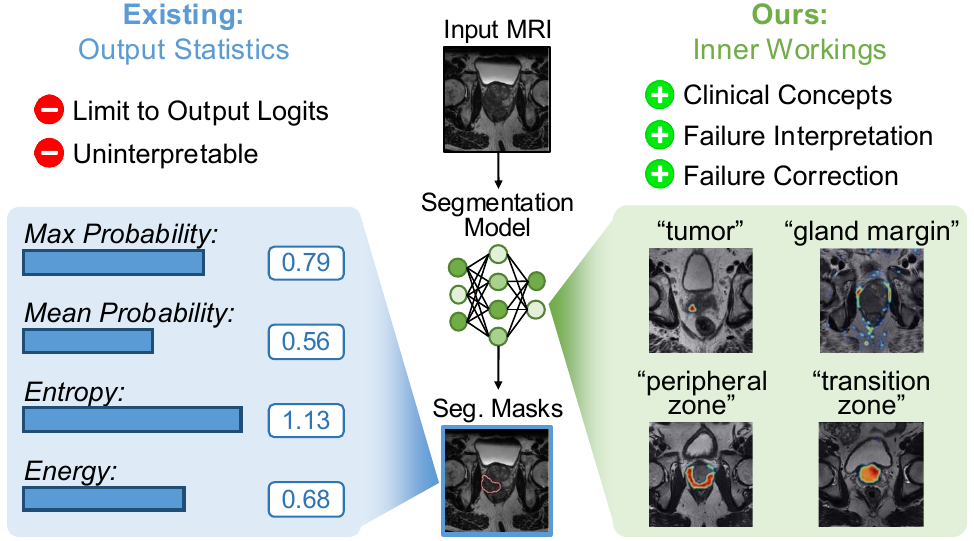}
  \caption{Two sources for capturing a model's \textit{Feeling of Error} (FOE). Output-level cues (left) are limited to uninterpretable logit statistics, offering no insight into failure causes. We instead extract the model's FOE from its inner workings (right), decomposing internal representations into interpretable concepts that enable failure detection, interpretation, and correction.}
  \label{fig:title}
  \vspace{-25pt}
\end{wrapfigure}
A straightforward way to approximate a model's Feeling of Error is through output-level cues~\cite{hendrycks2016baseline,thagaard2020can}, such as prediction confidence or uncertainty estimates (Fig.~\ref{fig:title}). However, this approach faces two fundamental limitations. First, these signals are rarely actionable. They may warn that a mask is risky but offer no insight into why a failure occurs. Second, correcting flagged failure using output-level cues faces a sensitivity–quality tradeoff: Increasing detection sensitivity catches more failures but also flags borderline, clinically acceptable cases, and unnecessary ``corrections'' can degrade overall segmentation quality. In our experiments, a 20-point gain in failure-detection F1-score via these proxies comes at the cost of a 8–10 point drop in segmentation DSC. This suggests that a model's feeling of error is likely encoded not in what it outputs, but in how it thinks (i.e., its decision-making process). 

\begin{figure*}[t]
  \centering
  \includegraphics[width=\linewidth]{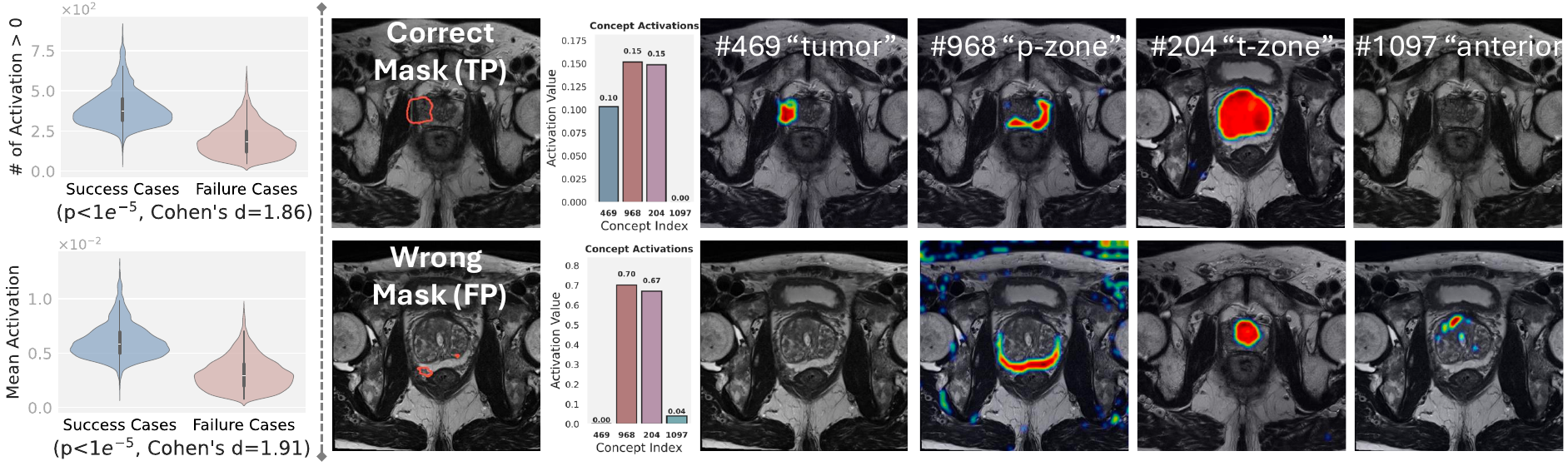}
  \caption{ Distributions of SAE concept activations for success and failure cases on PI-CAI. \textbf{Left:} Number of active concepts (activation value > 0) and Mean concept activation value. On average, {success cases have more active concepts with stronger activation values than failure cases}. Both metrics show statistically significant separation (p-value $< 1e^{-5}$, Cohen's d $> 1.8$), indicating that concept activation patterns distinguish correct from incorrect segmentation. \textbf{Right:} Example success (top) and failure (bottom) cases showing predicted masks and activation patterns for four key concepts. The bar charts display concept activation strengths for each case. Shared concepts \#968 (``peripheral zone'') and \#204 (``transition zone'') activate in both success and failure cases, reflecting anatomical context. However, {tumor-specific concept \#469 (``tumor'') shows strong activation only in the success case but minimal in the failure case}. \textit{These patterns suggest that the model encodes a ``feeling of error'' in its internal concept activations.}} 
  \label{fig:saestat}
\end{figure*}

To peek under the hood of the model’s decision-making, we employ mechanistic interpretability tools~\cite{bereska2024mechanistic,peng2026inside,sharkey2025open}, specifically Sparse Autoencoders (SAEs)~\cite{cunningham2023sparse,thasarathan2025universal,hindupur2025projecting}. These tools decompose high-dimensional neural activations into a dictionary of human-interpretable ``concepts.'' Our analysis show that ViT-based cancer segmentation models learn concepts that align with established clinical terms (Fig.~\ref{fig:title}). Crucially, we observe a distinct {latent signature} in failure cases: Incorrect segmentations activate fewer internal concepts with lower activation magnitudes compared to successful ones (Fig.~\ref{fig:saestat}, Left). The model, in some sense, ``knows'' it is wrong, even when its outputs say otherwise. This internal concept activation is the machine FOE we can leverage. Translating this signal for reliable failure detection raises two challenges. \textit{First}, \textit{where does the signal live?} Failures are not monolithic and may not localize to a single layer. A missed tumor might reflect an early-layer visual processing error~\cite{li2025bridging,daneshjou2022disparities}, while misclassifying benign tissue as malignant suggests a deep-layer semantic confusion~\cite{rayed2024deep}. \textit{Second}, \textit{how should the signal be modeled?} Failures may depend on interactions among concepts (and their spatial arrangement), where individually plausible cues become problematic only under certain co-activation patterns.

To tackle the challenges, we propose a two-stage framework. First, we build hierarchical ``failure representations'' by extracting concepts from early, middle, and deep layers. This gives us a comprehensive ``vocabulary'' for describing failures from low-level visual issues to high-level diagnostic confusions. Second, to enable both accurate failure detection and interpretation, we apply a lightweight classifier on these internal concepts. This classifier can learn complex interaction patterns (e.g., ``hyperintense + irregular texture + weak boundaries = likely failure'') while remaining interpretable: each concept's importance score shows exactly how a concept contributes to the failure prediction.

Our contributions: (1) We move beyond output logits to show that a model's internal concept representations can not only detect but also \textit{explain} segmentation failures in clinically meaningful terms. (2) We design a method that extracts failure-predictive concepts across network layers and pinpoints their diagnostic patterns using interpretable classifiers. (3) Empirical verification on prostate and pancreatic cancer shows that our approach achieves superior failure detection performance while providing intuitive explanations and maintaining high segmentation quality.

\section{Preliminaries and Findings}
\label{sec:related}
In this section, we begin by formally defining cancer segmentation failure, then present two empirical findings that support the existence of a \textit{machine FOE}, the model's internal signal that predicts segmentation failure.

\begin{figure*}[!thbp]
  \includegraphics[width=\linewidth]{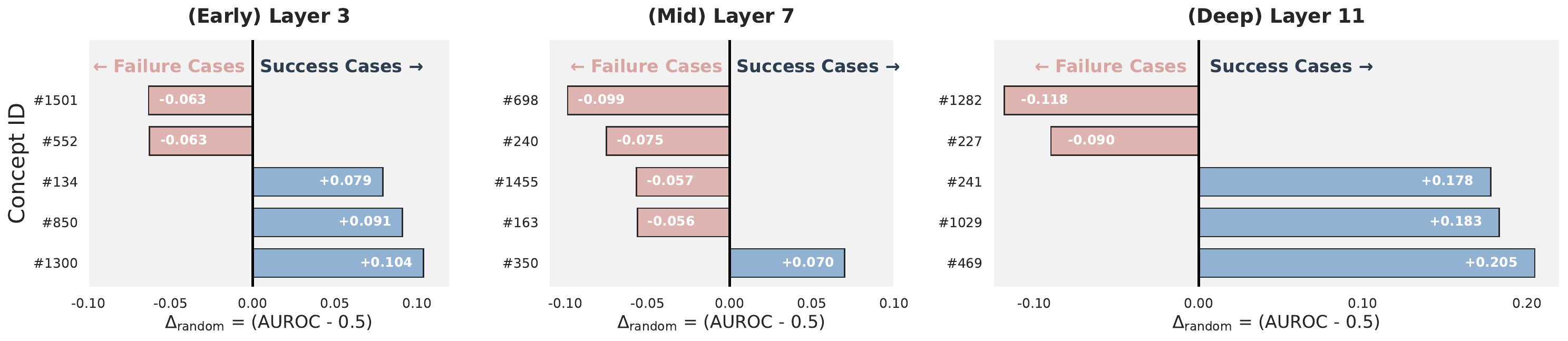}
  \caption{Predictive power of SAE concepts across layers for segmentation failures. We directly use each concept's activation score and the ground-truth failure labels to compute the AUROC for each concept. Each bar shows AUROC improvement over random chance ($\Delta_{\text{random}} = \text{AUROC} - 0.5$)  for the top 5 most discriminative concepts per layer. Blue bars indicate concepts predictive of success cases (success-predictive), while red bars indicate concepts predictive of failure cases (failure-predictive). \textit{Deep layers exhibit stronger predictive power overall, middle layers show more failure-predictive concepts, and early layers have weaker discriminative ability.}}
  \label{fig:localization}
\end{figure*}

\mydefinition{def:segmentation_failure}{
(Cancer Segmentation Failure). 
\textit{Let $f: \mathcal{X} \rightarrow \mathcal{M}$ be a segmentation model mapping medical images to binary masks, where $\mathcal{X}$ is the input image space and $\mathcal{M}$ is the mask space. For an image $x \in \mathcal{X}$, let $\hat{m} = f(x)$ denote the predicted mask and $m^* \in \mathcal{M}$ the ground truth mask. Let $\mathcal{E}: \mathcal{M} \times \mathcal{M} \rightarrow [0,1]$ be a quality metric (e.g., Dice coefficient). A \textbf{cancer segmentation failure} occurs when $\mathcal{E}(\hat{m}, m^*) \le \tau$ for a predefined threshold $\tau \in [0,1]$.}
}

\textbf{SAE preliminary.} SAEs learn interpretable, overcomplete representations by decomposing inputs into sparse combinations of learned features~\cite{bricken2023towards,ng2011sparse,gao2024scaling,li2026inside,chen2026world}. Given an internal activation $\mathbf{h} \in \mathbb{R}^d$ from a ViT layer; The SAE consists of an encoder $E: \mathbb{R}^{d} \to \mathbb{R}^{D}$ (where $D > d$) and decoder $G: \mathbb{R}^{D} \to \mathbb{R}^{d}$:
\begin{equation}
\begin{aligned}
& \mathbf{z} = E(\mathbf{h}) = \text{ReLU}(\mathbf{W}_{\text{enc}} \mathbf{h} + \mathbf{b}_{\text{enc}}), \quad \\ &\hat{\mathbf{h}} = G(\mathbf{z}) = \mathbf{W}_{\text{dec}} \mathbf{z} + \mathbf{b}_{\text{dec}},
\end{aligned}
\label{eqn:sae}
\end{equation}
where $\mathbf{W}_{\text{enc}}, \mathbf{W}_{\text{dec}}^{\top} \in \mathbb{R}^{D \times d}$ and $\mathbf{b}_{\text{enc}}, \mathbf{b}_{\text{dec}} \in \mathbb{R}^{D}$. The sparse latent $\mathbf{z} \in \mathbb{R}^{D}$ represents concept activations, where each dimension $z_i$ corresponds to a learned feature. The decoder reconstructs inputs as $\hat{\mathbf{h}} = \sum_{i=1}^{D} z_i \mathbf{w}_i + \mathbf{b}_{\text{dec}}$, where $\mathbf{w}_i$ (the $i$-th column of $\mathbf{W}_{\text{dec}}$) is the $i$-th feature direction. SAEs are trained to minimize: $\mathcal{L}(\mathbf{h}) = \|\mathbf{h} - \hat{\mathbf{h}}\|_2^2 + \lambda \|\mathbf{z}\|_1$, balancing reconstruction fidelity with sparsity.

We train SAEs on internal activations of ViT-based segmentation models (e.g., Medical SAM variants~\cite{ma2025rationale,ma2024medsam,chen2024ma}) fine-tuned across multiple cancer datasets. This yields two key findings.

\textbf{Finding 1: Failure cases exhibit distinct internal concept activation patterns.} As shown in Fig.~\ref{fig:saestat}, failure cases activate significantly fewer SAE concepts and at substantially lower magnitudes compared to successful cases ($p < 10^{-5}$, Cohen's $d = 1.8$). This large effect size suggests that failures are not merely low-confidence predictions but correspond to a qualitatively different internal state. To further quantify the discriminative power of individual concepts, we compute the AUROC of each concept's activation value against ground-truth failure labels. As shown in Fig.~\ref{fig:localization}, this reveals two distinct groups: \textit{success-predictive} concepts (high AUROC, strongly active in successful cases) and \textit{failure-predictive} concepts (disproportionately elevated in failure cases). The existence of failure-predictive concepts suggests that failures leave a detectable signature in the latent space.

\textbf{Finding 2: There are SAE concepts that align with established clinical terms.} Beyond their discriminative value, we find that SAE concepts correspond to clinically meaningful anatomical structures (Figs.~\ref{fig:saestat} and~\ref{fig:method}). To validate this, we follow a two-stage protocol. First, we automatically measure spatial overlap between each concept's activated pixels and annotated lesion or anatomical masks across all datasets: concepts with IoU $\geq 0.5$ are assigned clinical labels based on the corresponding annotation. We will introduce how to obtain the pixels corresponding to SAE concepts in the next section. Second, we present a representative set of automatically verified concept-label pairs, along with their pixel localizations, to two board-certified radiologists, who independently assess whether each label accurately describes the visualized concept. Both radiologists confirmed agreement with all presented labels. Together, these two findings establish that SAE concepts are both \textit{semantically grounded} and \textit{failure-informative}.

\textbf{Open problems.} While these findings are promising, two critical questions remain. (1) How to effectively identify concepts that are predictive of failures? The SAE concept dictionary can reach 1K–10K entries, making manual inspection or heuristic selection infeasible. (2) Even after discriminative concepts are obtained, how to best use them for failure detection remains an open problem. Segmentation models distribute information across layers; it is unclear which layers’ concepts are most informative for detecting failures, or how to effectively aggregate them into a reliable decision signal. In the next section, we address these challenges to build a robust failure detector.

\section{Detecting Failures via Internal Concepts}
\label{sec:method}

\begin{figure*}[!thbp] 
  \includegraphics[width=1.0\linewidth]{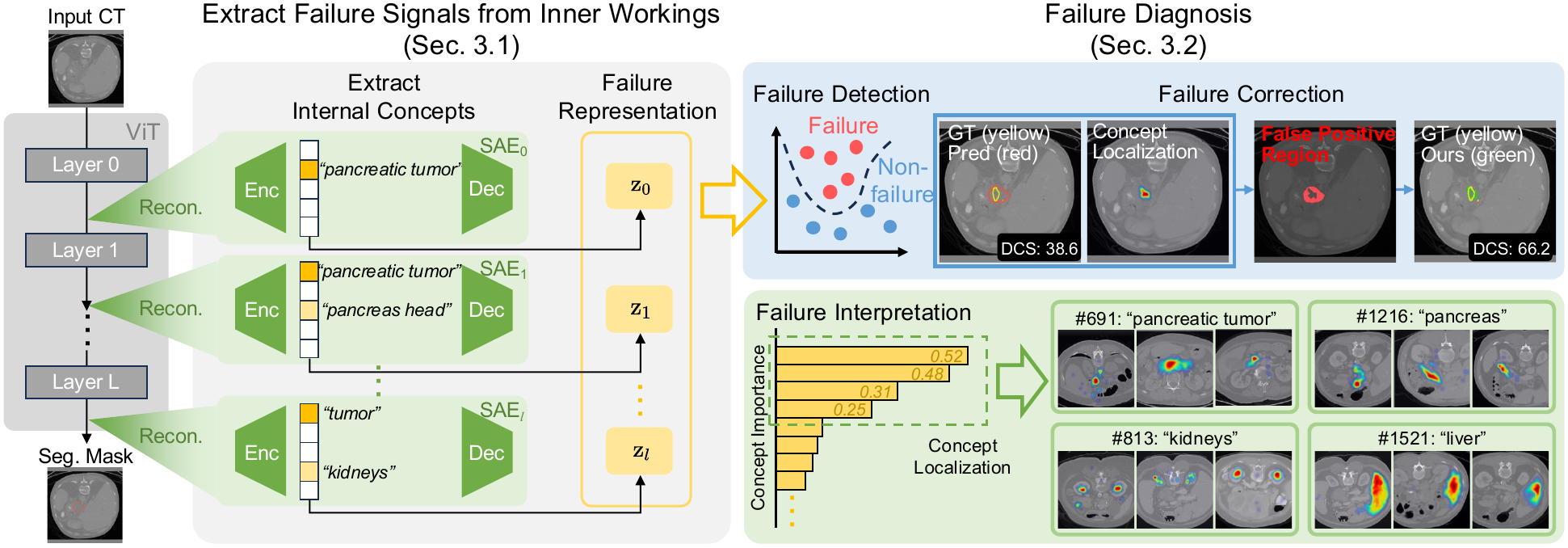}
  \caption{{Overview of the proposed framework.} \textit{{Failure signal extraction}}: SAEs decompose internal representations from multiple layers of a frozen segmentation model into sparse concept vectors, capturing failure-relevant signals across the full representational hierarchy. \textit{{Failure diagnosis}}: The concatenated concept representation is passed to a lightweight classifier that serves dual purposes: \textit{detection}, which identifies and corrects segmentation failures, and \textit{interpretation}, which ranks concepts by their contribution to the failure prediction, tracing model errors back to interpretable concepts}
  \label{fig:method}
\end{figure*}

This section details our framework that leverages segmentation models' internal concepts for failure detection. We address two key questions: (1) \textit{How to capture all potential failure signals?} Rather than focusing on a single ``best'' layer, we capture all possible signals of failure from all layers, obtaining the holistic representation of failures (Sec.~\ref{sec:signal}). (2) \textit{How to identify predictive patterns of failure?} Failures may not be linked to a single concept, but to interactions between them. We build a dedicated classifier on top of failure representation to automatically learn the ``signature of failure'' (Sec.~\ref{sec:classifier}). Fig.~\ref{fig:method} illustrates our complete pipeline.

\subsection{How to Capture All Potential Failure Signals?}
\label{sec:signal}

Cancer segmentation failures could arise from diverse clinical causes that manifest across different representational levels. Low-level image artifacts (motion, noise, poor contrast) create visual anomalies, while high-level anatomical complexities (ambiguous boundaries, irregular morphology) cause semantic confusion. Thus, a robust detector should analyze the model's representations from low-level visual features to high-level semantics to holistically characterize all potential failure signals.

\textbf{Method.} To obtain representations of model failures using internal concepts, we train SAEs on latent embedding from early, mid, and deep layers of the segmentation model $f$. Specifically, we uniformly sampled across the ViT backbone from layers $L = \{1, 3, 5, 7, 9, 11\}$ to ensure coverage of the full representational spectrum and extract their latent embeddings $\mathbf{h}_l(x) \in \mathbb{R}^{d}$ for each input image $x$ and layer $l \in L$. $d$ is the latent embedding size. For each layer, we train an independent SAE to transform dense embeddings into sparse, interpretable concept vectors. The SAE encoder $E_l: \mathbb{R}^{d} \rightarrow \mathbb{R}^D$ is trained on embeddings from both correct predictions and failures $\{ \mathbf{h}_l(x) \mid x \in \mathcal{X}_{C} \cup \mathcal{X}_{F} \}$, where $D$ denotes the dictionary size. Each encoder produces sparse concept activations $\mathbf{z}_l(x) = E_l(\mathbf{h}_l(x)) \in \mathbb{R}^D$.
We obtain the holistic failure representation by concatenating all sparse concept vectors across layers: 
\begin{equation}
\mathbf{c}(x) = {\textstyle \bigoplus_{l \in L}} \mathbf{z}_l(x) = {\textstyle \bigoplus_{l \in L}} E_l(\mathbf{h}_l(x)) \in \mathbb{R}^{|L| \cdot D},
\end{equation}
where $\bigoplus$ denotes concatenation and $|L| = 6$ is the number of selected layers. $\mathbf{c}(x)$ preserves all layer-specific internal concepts, providing a rich signal for failures.

\textbf{Validation of discriminability.} Before proceeding to the next section, we first verify that the learned representation $\mathbf{c}(x)$ captures discriminative information about failures. Fig.~\ref{fig:saestat} visualizes the distribution of failure representations for correct prediction and failures. The two classes exhibit statistically significant separation with minimal overlap (independent t-test, p $< 1e^{-5}$ and effective size, d $> 1.8$). This empirical evidence confirms that internal concepts encode discriminative features suitable for failure detection.

\subsection{How to Identify Predictive Patterns of Failure?}
\label{sec:classifier}

Given the failure representation $\mathbf{c}(x) \in \mathbb{R}^{|L| \cdot D}$, our goal is to find a decision function $g: \mathbb{R}^{|L| \cdot D} \rightarrow [0,1]$ that satisfies two objectives: (1) \textit{accurate prediction}, where $p_{\text{failure}}(x) = g(\mathbf{c}(x))$ achieves high discriminative for failure prediction, and (2) \textit{interpretability}, where $g$ provides explainable feature importance scores $\mathbf{w} \in \mathbb{R}^{|L| \cdot D}$ that quantify each concept's contribution to failure detection.

\textbf{Method.} We learn the decision function $g$ in a data-driven manner. The first step is to construct a labeled \textit{failure dataset} $\mathcal{D} = \{(\mathbf{c}_i, y_i)\}_{i=1}^{N}$, where $\mathbf{c}_i = \mathbf{c}(x_i) \in \mathbb{R}^{|L| \cdot D}$ is the failure representation for image $x_i$, and $y_i \in \{0,1\}$ indicates correct prediction ($y_i=0$) or failure ($y_i=1$). The dataset $\mathcal{D}$ is derived from the segmentation model's training and validation sets, where ground truth annotations enable reliable failure labeling. For both accuracy and interpretability, we model $g$ as a linear or non-linear classifier (e.g., XGBoost~\cite{chen2016xgboost} or logistic regression). The classifier is trained on the failure dataset, optimizing binary cross-entropy loss:  
\begin{equation}
\begin{aligned}
\min_{g} \mathcal{L}(g, \mathcal{D}) = & -\frac{1}{N} \sum_{i=1}^{N} \Big[ y_i \log g(\mathbf{c}_i) + (1 - y_i) \log(1 - g(\mathbf{c}_i)) \Big] + \lambda \Omega(g),
\end{aligned}
\label{eqn:cls}
\end{equation}
where $\Omega(g)$ controls model complexity, and $\lambda \geq 0$ is the regularization strength.

\textbf{Classifier and feature importance.} We adopt XGBoost~\cite{chen2016xgboost} as our primary classifier. Gradient boosting is particularly well-suited for this task as it: naturally handles high-dimensional sparse features, learns non-linear interactions between concepts across layers, and provides feature importance scores that reveal which concepts are most discriminative for failure detection. We also benchmark against other classifiers in the Ablation Study (Tab.~\ref{ablation:classifier}). Once trained, the classifier provides importance scores for each concept dimension in $\mathbf{c}(x)$, quantifying their contribution to failure prediction. By mapping these scores back to specific SAE concepts and their corresponding layers, we can identify which internal concepts, and at what representational levels, are most indicative of segmentation failures. Fig.~\ref{fig:importance} visualizes the learned most important concepts.

\subsection{Implementation and Discussion}

\textbf{Patch- vs. image-level embedding}. We adopt ViTs as our backbone for cancer segmentation. A fundamental design choice when applying SAEs to ViTs is: should we aggregate patch tokens into a single image-level representation, or preserve spatial structure by training SAEs directly on patch-level embeddings? Prior work mainly focuses on the classification task and predominantly uses image-level aggregation~\cite{DasMuh_CytoSAE_MICCAI2025,renzulli2025medsae,le2024learning}, where patch embeddings are pooled into a global embedding before SAE training. This is sufficient for classification, as global semantic concepts are adequate for category prediction. However, segmentation fundamentally differs: it is a \textit{dense prediction task} requiring pixel-wise decisions. Therefore, we train all SAEs directly on patch-level embeddings to preserve spatial granularity. 

\textbf{Concept interpretation via spatial localization.} To understand what each concept represents, we leverage the spatial structure preserved in patch-level representations. Given an input image $x$, the ViT backbone produces spatial patch embeddings $\mathbf{H}_l(x) \in \mathbb{R}^{h \times w \times d}$ at layer $l$, where $h$ and $w$ denote the spatial grid dimensions. Applying the SAE encoder $E_l$ to each spatial location yields patch-level concept activations:
\begin{equation}
\mathbf{C}_l(x) = E_l(\mathbf{H}_l(x)) \in \mathbb{R}^{h \times w \times D},
\end{equation}
where $\mathbf{C}_l(x)[i,j,:] \in \mathbb{R}^{D}$ represents the concept vector at spatial position $(i,j)$. To visualize where a specific concept $k$ activates in the image, we extract its spatial activation map: $\mathbf{C}_l^{(k)}(x) = \mathbf{C}_l(x)[:,:,k] \in \mathbb{R}^{h \times w}$. This 2D map reveals which image regions mostly activate concept $k$.

\textbf{Failure correction.} Our framework enables principled \textit{failure correction} as a natural extension of detection. Since the classifier operates at the patch level, we can identify not just whether a prediction is a false positive, but which specific patches drive the wrong decision. We formalize this as a patch retention problem: the corrected segmentation retains patch $(i,j)$ if and only if its local failure score falls below a threshold $\eta$:
\begin{equation}
\hat{m}_{\text{corrected}}(i,j) = \hat{m}(i,j) \cdot \mathbf{1}\!\left[g\!\left(\mathbf{c}(x)_{ij}\right) < \eta\right],
\label{eqn:correction}
\end{equation}
where $\mathbf{c}(x)_{ij}$ is the concept vector at patch $(i,j)$. This selective correction preserves high-confidence TP regions that output-based methods would discard entirely when flagging a prediction as failed, improving both precision and the clinical utility of the corrected mask.

\textbf{Relation to broader context.} Our framework shares the high-level ``concept-then-predict'' motivation with concept bottleneck models~\cite{koh2020concept}, but differs in two critical ways: (1) it operates post-hoc on any pre-trained segmentation model without architectural modifications, and (2) it discovers concepts unsupervised via SAEs, eliminating the need for manual concept annotations. Additionally, recent medical vision-language models can map visual embeddings to discrete vocabularies aligned with language space~\cite{kim2024transparent,gao2024aligning,hou2024self}. By contrast, our approach does not require internal concepts to be fully aligned with human language. In fact, some of the most predictive failure concepts we discover are non-semantic. They capture internal processing patterns that correlate with failures but are not human-readable. This aligns with recent arguments that AI systems may develop representations outside our existing vocabulary~\cite{hewitt2025we}.

\section{Experiments}
\label{sec:experiment}

\subsection{Data, Baselines, Metrics, and Hyperparameters}

\textbf{Datasets.} We evaluate our methods on three public cancer datasets: Prostate (PI-CAI~\cite{saha2024picai}, Prostae158~\cite{adams2022prostate158}) and Pancreatic (PanTS~\cite{li2025pants}). \textit{PI-CAI} contains 1,500 mp-MRI scans from three Dutch medical centers. We used its provided annotations for prostate lesions and partitioned the dataset into 1,200 training, 60 validation, and 240 testing scans. PI-CAI is one of the largest publicly available MRI datasets for prostate cancer. \textit{Prostate158} includes 158 mp-MRIs with T2W,  DWI and ADC sequences. We use all data for zero-shot evaluation. \textit{PanTS} consists of 9,901 CT scans with annotations for the pancreatic tumors, pancreas, and its head, body, and tail. We use the subset PanTS to curate our data. Specifically, we randomly selected 700 CTs and split them into 448 training, 112 validation, and 140 testing scans.

\textbf{Baselines.} We compare against output-based uncertainty quantification methods that do not require additional training: maximum softmax probability (MaxProb)~\cite{hendrycks2016baseline}, mean softmax probability (MeanProb)~\cite{hendrycks2016baseline}, entropy~\cite{hendrycks2016baseline}, energy score~\cite{liu2020energy}, and direct probability thresholding. These methods represent the standard practice of using model confidence as a proxy for prediction quality. We do not compare against approaches that incorporate failure detection into the training objective, as such methods require access to and modification of the main network's training pipeline, which is incompatible with the post-hoc deployment scenario this work targets. In clinical practice, segmentation models are typically received as pre-trained, frozen systems from the vendor; our setting and all baselines evaluated here reflect this realistic constraint.

\textbf{Evaluation metrics.} We evaluate segmentation performance using the Dice Similarity Coefficient (DSC, $\uparrow$). For failure detection performance, we report the F1 ($\uparrow$) score as our primary metric. Additionally, following common practice~\cite{liu2020energy,zhao2025verifying,nguyen2025interpretable}, we report the AUROC ($\uparrow$) to assess how well the method ranks correct versus incorrect segmentations, and the True Positive Rate at 95\% False Positive Rate (FPR95, $\downarrow$) to measure reliability under strict conditions.

\textbf{Hyperparameter settings.} Our segmentation backbone is MedSAM~\cite{ma2024medsam}, a widely-adopted foundation model with an image encoder, prompt encoder, and mask decoder. To enable automatic segmentation, we modify the prompt encoder to operate without manual prompts~\cite{shaharabany2023autosam}. We fine-tune the model for each cancer segmentation task on its corresponding dataset using Focal Loss~\cite{lin2017focal} ($\alpha=0.97$, $\gamma=2$) and the Adam optimizer~\cite{kingma2014adam} with a $1e^{-4}$ weight decay and an exponential learning rate scheduler. For SAE training, we follow the BatchTopK~\cite{bussmann2024batchtopk} approach with learning rate $1e^{-4}$ for 5 epochs. We use dictionary size $D=1,536$ and sparsity $S=8$ across all layers.

\subsection{Failure Interpretation}
\textit{\textbf{Takeaway:} The model's internal concepts provide a mechanistic explanation for segmentation failures and a reliable signal for predicting them.}

\textbf{Model's internal concepts.} To understand what the segmentation model learns, we train SAEs on its latent embeddings. The SAEs disentangle the polysemantic embeddings into monosemantic ones, which can be treated as the internal concepts. As shown in Figs.~\ref{fig:title} and~\ref{fig:saestat}, these concepts localize to anatomically relevant regions. In prostate cancer, for instance, we identify distinct concepts for the peripheral zone, gland, and tumor, with finer-grained concepts distinguishing tumor core from edge. We observe that a similar anatomical alignment appears consistently in pancreatic cancer.

\textbf{Failure analysis.} Given these internal concepts, we next analyze how they behave when segmentation fails. We find that the activation of these concepts can distinguish success from failure cases. In Fig.~\ref{fig:saestat}, successful ones produce strong activations in tumor concepts. In contrast, failures activate these concepts weakly or not at all, suggesting the model internally ``knows'' it is uncertain about these areas. This provides a mechanistic explanation for failures: failures often occur when the model misinterprets ambiguous image features, failing to activate the correct internal representations.

\begin{figure*}[t]
  \centering
  \includegraphics[width=\linewidth]{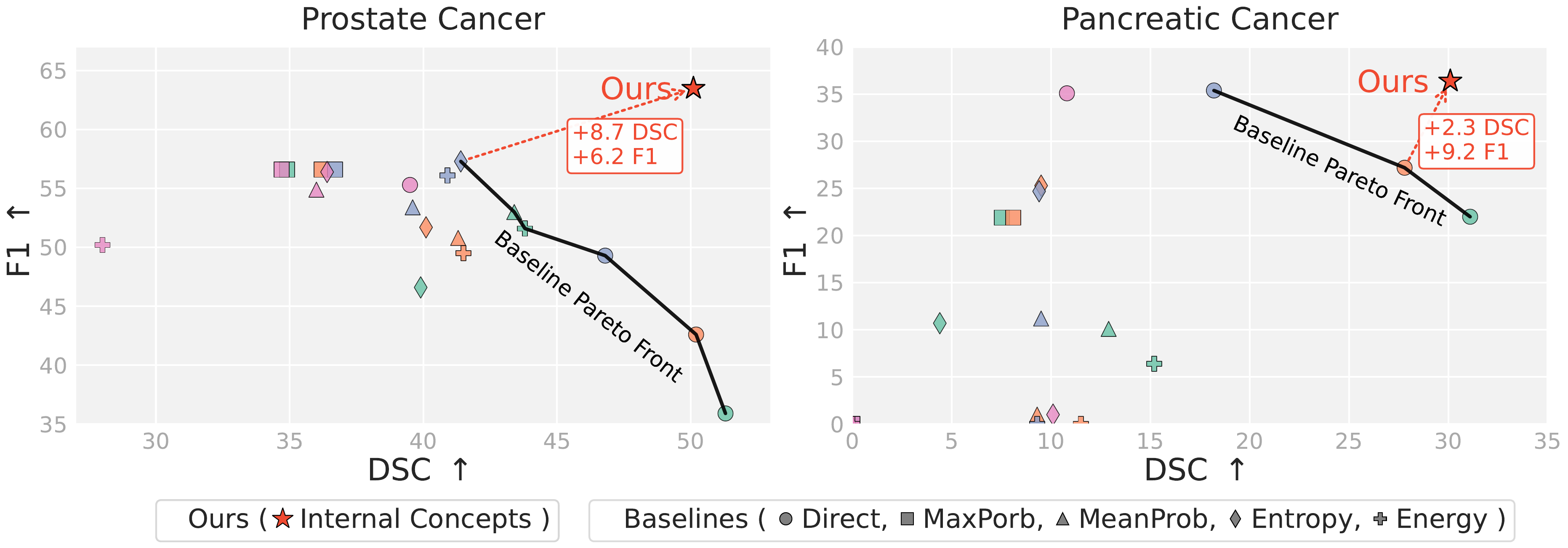}
  \caption{F1 ($\uparrow$) vs. DSC ($\uparrow$) scores on PI-CAI (Prostate Cancer) and PanTS (Pancreatic Cancer). Each marker represents a baseline method varying by threshold [0.5, 0.8]. The solid gray line indicates the Pareto front, showing the best achievable trade-off between segmentation accuracy (DSC) and failure detection quality (F1). \textit{Our method surpasses the baseline Pareto front, achieving both higher DSC and higher F1, demonstrating superior reliability without sacrificing segmentation quality.}}
  \label{fig:detection}
\end{figure*}

\subsection{Failure Detection}
\textit{\textbf{Takeaway:} Model's internal concepts contain richer signals about failures.}

Our analysis has shown that the internal concept activations are discriminative for segmentation failures. Do internal concepts provide more discriminative information for identifying failures than the model's final output confidence? To answer this question, we train a simple classifier using these internal concept activations to predict segmentation failures and compare it against baselines that use only the model's output (e.g., logits). The results in Fig.~\ref{fig:detection} and Tab.~\ref{tab:zeroshot} show that our models are robust and generalizable.

\textbf{Quantitative performance.} We evaluate our method across three cancer segmentation datasets. The results are shown in Fig.~\ref{fig:detection}. Baseline methods exhibit a clear Pareto tradeoff between segmentation quality (DSC) and failure detection performance (F1); 
Improving one metric requires sacrificing the other. By leveraging internal concepts, our method breaks this tradeoff, simultaneously achieving high performance on both metrics.

\begin{figure}[t]
  \centering
  \includegraphics[width=0.9\linewidth]{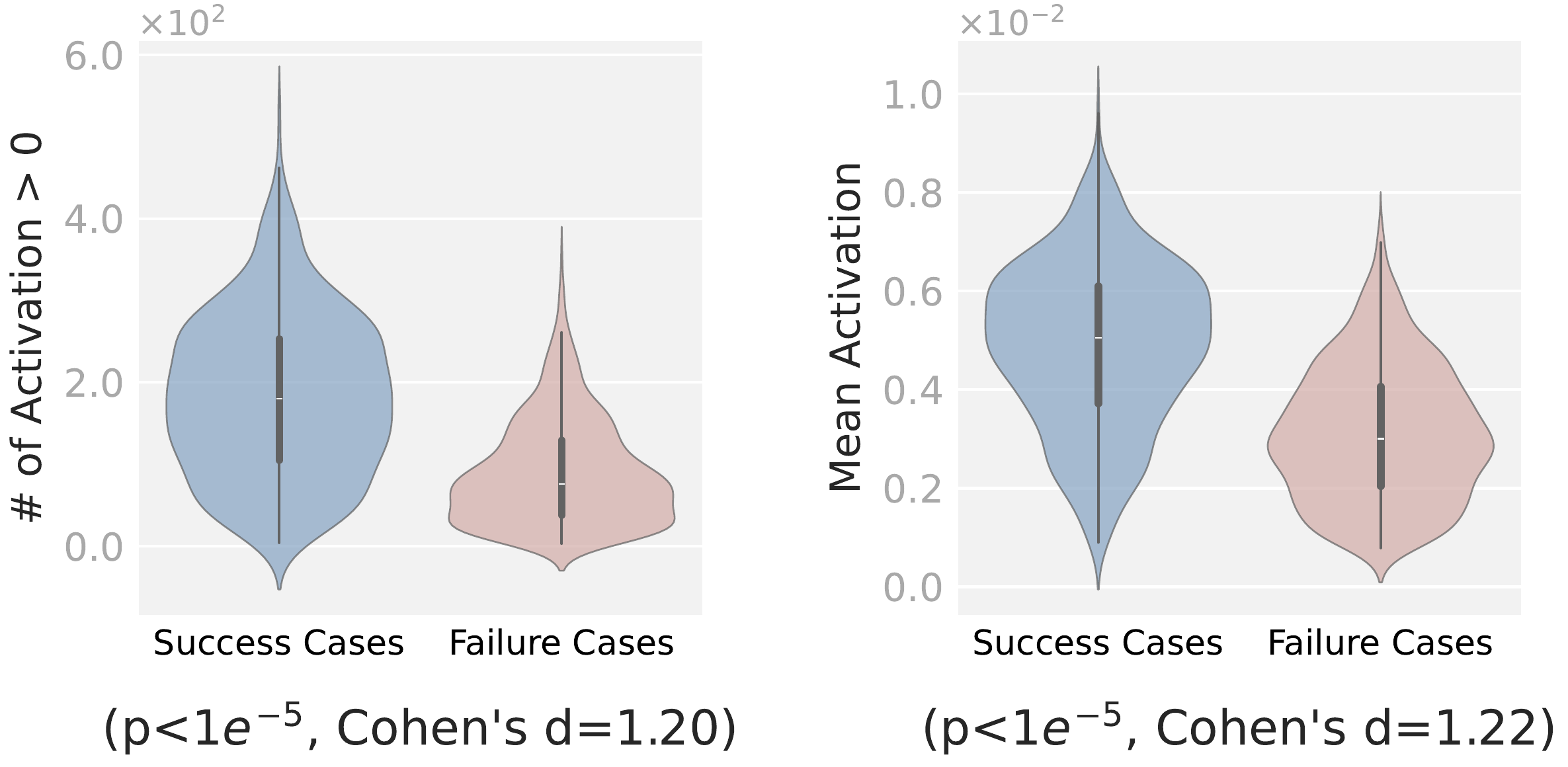}
  \caption{ SAE concept activations for success and failure cases on Prostate158 (unseen data). Both metrics show statistically significant separation (p $< 1e^{-5}$, Cohen's d $> 1.2$), indicating that \textit{concept activation patterns can generalize across datasets}.}
  \label{fig:sae158}
\end{figure}

\setlength{\intextsep}{2pt}%
\setlength{\columnsep}{10pt}%
\begin{wraptable}{r}{0.55\linewidth}
    \centering
    \vspace{-11.5pt}
    \caption{Zero-shot failure detection on the unseen Prostate158 dataset using detectors trained on PI-CAI. \textit{Our concept-based detector consistently outperforms confidence-based baselines across all metrics.}}

\label{tab:zeroshot}
\resizebox{\linewidth}{!}{%
\begin{tabular}{lccc}
\toprule
              & FPR95 ($\downarrow$) & AUROC ($\uparrow$) & AUPR ($\uparrow$) \\ \midrule
MaxProb       & 100.0                & 50.0               & 64.4              \\
MeanProb      & 100.0                & 52.6               & 65.1              \\
Entropy       & 97.8                 & 53.1               & 65.6              \\
Energy        & 100.0                & 48.0               & 63.6              \\ \midrule
\textbf{Ours} & \textbf{64.4}        & \textbf{73.6}      & \textbf{74.1}     \\ \bottomrule
\end{tabular}}

\end{wraptable}
\textbf{Zero-shot generalization.}
We evaluate all detectors trained on PI-CAI directly on the unseen Prostate158 dataset without fine-tuning. As shown in Tab.~\ref{tab:zeroshot}, confidence-based baselines, such as MaxProb and Entropy, generalize poorly, with AUROC scores close to random chance. 
In contrast, our concept-based detector maintains substantially better performance across all three metrics. This indicates that internal concepts are more reliable and generalizable features for failure detection than model confidence, which can be poorly calibrated on unseen data. Fig.~\ref{fig:sae158} further validates generalization: on the unseen Prostate158 dataset, success and failure cases remain separable in concept space, demonstrating that learned failure patterns transfer across domains.

\begin{figure}[t]
  \includegraphics[width=\linewidth]{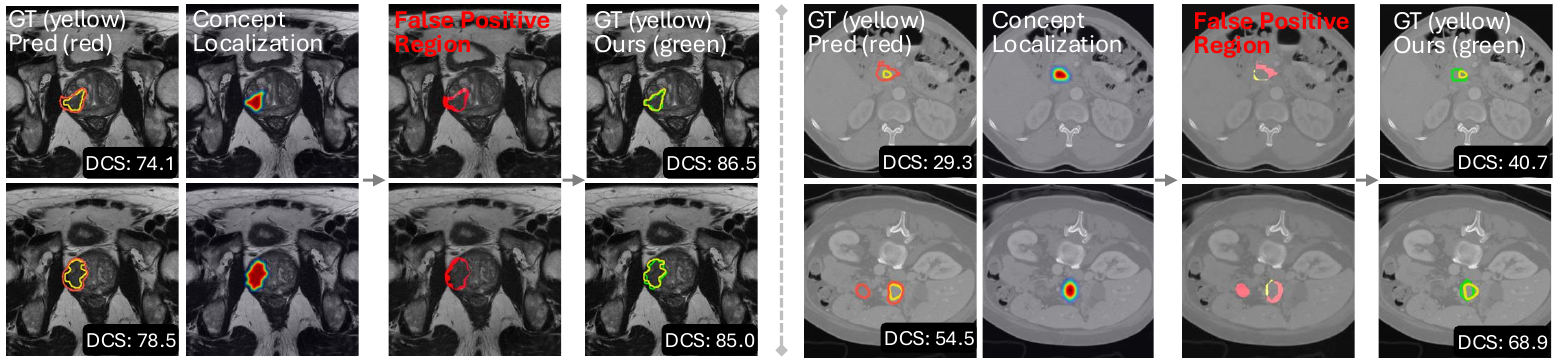}
  \caption{Visualization of Failure correction on PI-CAI (MRI, left) and PanTS (CT, right). For each example, four columns are shown: (\textit{1st}) initial segmentation prediction (red) and ground truth (yellow); (\textit{2nd}) SAE concept localization; (\textit{3rd}) detected false positive regions (red);  (\textit{4th}) final prediction mask (green) after FP removal. DCS scores after correction demonstrate consistent improvement across both datasets. \textit{SAE concepts enable the localization and removal of false-positive regions, consistently improving segmentation accuracy across both MRI and CT datasets.}}
  \label{fig:tpfpmask}
\end{figure}

\textbf{Qualitative analysis of detected failures.} Fig.~\ref{fig:tpfpmask} shows representative failures detected by our method: (1) Hallucinated lesions where the model predicts non-existent tumors, and (2) Boundary errors where true lesions are localized but severely over- or under-segmentation. Our approach identifies both complete false positives and boundary-level errors.

\begin{figure}[!t]
\centering
   \includegraphics[width=0.98\linewidth]{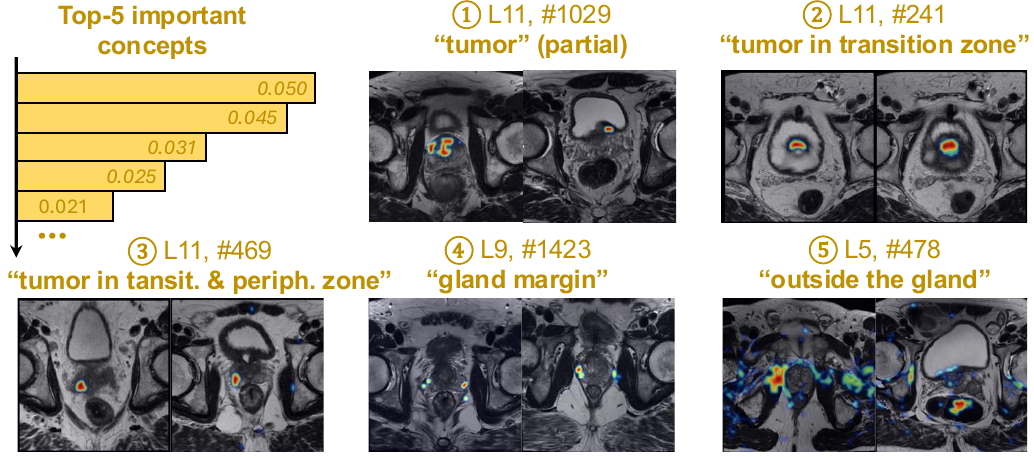} 
  \caption{Top-5 concept importance for failure prediction. Concepts are ranked by importance scores and visualized via localization heatmaps. Tumor-related concepts (1-3) dominate the ranking, capturing partial tumors, transition zone (TZ) tumors, and tumors spanning TZ and peripheral zone (PZ). Anatomical boundary concepts (4-5) identify gland edges and extra-prostatic regions, \textit{revealing that the classifier leverages both tumor characteristics and anatomical context for failure detection.}}
  \label{fig:importance}
\end{figure}

\textbf{Analysis of concept importance.} Since our classifier is trained on using the internal concepts, we can obtain each concept's importance. Fig.~\ref{fig:importance} visualizes the concept importance scores and concept localization. Crucially, we observe that the most important concepts are tumor-related.

\subsection{Ablation Study}

\textbf{SAE architecture.} We ablate SAE sparsity and dictionary size to determine their impact on failure detection performance. Results are shown in Tab.~\ref{tab:ablation_sae}. Optimal performance occurs at sparsity L0 $=$ 8 and dictionary size $=$ 1,536, which are notably lower than typical values for natural images (S $\approx$ 16, dictionary size D $\approx$ 10,000)~\cite{cunningham2023sparse,thasarathan2025universal}. We hypothesize this reflects the more structured nature of medical imaging modalities compared to natural images.

\textbf{Early vs. deep layer.} We analyze the contribution of concepts from different network depths (Tab.~\ref{tab:layers}). Deep layer concepts achieve the highest failure detection performance (F1=57.6) by capturing high-level semantic patterns, while mid-layer concepts excel at segmentation quality (DSC=48.8). Early layer concepts, encoding low-level visual features, contribute moderately to both tasks.

\textbf{Choice of failure classifier.} We tried different classifiers such as logistic regression, random forest, and XGB classifier. The results are shown in Tab.~\ref{ablation:classifier}. The XGB classifier gives the best results.

\textbf{Computation overhead.} Tab.~\ref{tab:computation} shows the per-image latency on a single RTX A6000 GPU at $1024^2$ input (100 forwards, mean$\pm$std). The decoder overhead specifically is $\approx 3.4$ ms and is invoked \textit{only} at intervention time, so routine deployment adds only $+3.7\%$. Memory overhead is also small: $+60$ MB GPU memory ($+1.8\%$) and 54 MB disk storage for all six SAE checkpoints. Relative overhead is stable across batch sizes $1$--$16$ ($\Delta$ encoder $3.5$--$4.3\%$, $\Delta$ enc+dec $6.5$--$7.4\%$). Our full failure detection pipeline runs at around $5.66$ img/s.

\begin{table}[t]
\centering
\begin{minipage}[t]{0.48\linewidth}
    \centering
\caption{Ablation study on SAE architecture. We vary the dictionary size (D) and sparsity (S) and report the resulting failure detection F1 and segmentation DSC. The optimal configuration (D=1536, S=8) is in bold.}
\label{tab:sae_ablation}
\resizebox{\columnwidth}{!}{
\begin{tabular}{l ccc ccc}
\toprule
 & \multicolumn{3}{c}{\text{D=1,536 (768$\times$2)}} & \multicolumn{3}{c}{\text{D=3,072 (768$\times$4)}} \\
\cmidrule(lr){2-4} \cmidrule(lr){5-7}
\text{Sparsity (S)} & 4 & 8 & 16 & 4 & 8 & 16 \\
\midrule[1pt]
DSC ($\uparrow$) & 49.7 & \textbf{50.1} & 49.9 & 50.1 & 49.7 & 49.6 \\
F1 ($\uparrow$)  & 60.3 & \textbf{63.5} & 62.5 & 58.4 & 57.4 & 62.6 \\
\bottomrule
\end{tabular}}
\label{tab:ablation_sae}
\end{minipage}
\hfill
\begin{minipage}[t]{0.48\linewidth}
\centering
\caption{Ablation on concept source depth. We report results using concepts from early, middle, and deep layers. \textit{Combining all three levels (All) achieves the best DSC and F1}, indicating that concepts across network depths provide complementary information.}
\resizebox{\linewidth}{!}{%
\begin{tabular}{lccc|c}
\toprule
\multicolumn{1}{c}{Layers (L)} &  Early (1, 3) & Mid (5, 7) & Deep (9, 11)                 & All  \\ \midrule
DSC ($\uparrow$)                   &  47.5     &  48.8   & \multicolumn{1}{c|}{49.6} & \textbf{50.1} \\
F1 ($\uparrow$)                    &  44.5     &  50.4   & \multicolumn{1}{c|}{57.6} & \textbf{63.5} \\ \bottomrule
\end{tabular}}
\label{tab:layers}
\end{minipage}
\end{table}

\begin{table}[t]
\centering
\begin{minipage}[t]{0.48\linewidth}
\centering
\caption{Ablation on the failure detection classifier. We compare performance using our SAE concepts with different classifiers. \textit{XGBoost achieves the best results for both DSC and F1.}}
\resizebox{\linewidth}{!}{%
\begin{tabular}{lcccc}
\toprule
Classifier    & XGB & Logistic & Decision Tree & Random Forest \\ \midrule[1pt]
DSC ($\uparrow$) &  \textbf{50.1}     &  47.8   &  48.5    &  44.6   \\
F1 ($\uparrow$)  &  \textbf{63.5}     &  59.2   &   58.7   &  57.0   \\ 
\bottomrule
\end{tabular}}
\label{ablation:classifier}
\end{minipage}
\hfill
\begin{minipage}[t]{0.48\linewidth}
\centering
\caption{Computation overhead. We compute the per-image latency for our failure detection method. \textit{Our method introduce limited runtime overhead.}}
\resizebox{\linewidth}{!}{
\begin{tabular}{l|lc}
\toprule
Configuration & Latency (ms) & $\Delta$ \\ \midrule
ViT model (e.g., MedSAM) & $126.3\pm0.5$ & --- \\ 
+ 6 SAE encoders (routine) & $130.9\pm2.7$ & $+3.7\%$ \\
+ 6 SAE enc+dec (intervention) & $134.3\pm0.3$ & $+6.3\%$ \\
+ XGBoost (full pipeline) & $176.5\pm13.3$ & $+39.7\%$ \\
\bottomrule
\end{tabular}
}
\label{tab:computation}
\end{minipage}
\end{table}

\section{Related Work}

\textbf{Medical image segmentation for cancer detection.} Deep learning has revolutionized cancer segmentation in medical imaging, with U-Net~\cite{ronneberger2015u} and its variants~\cite{isensee2021nnu} becoming the de facto standard for tumor delineation across multiple modalities including MRI, CT, and ultrasound. Recent advances have focused on improving segmentation accuracy through architectural innovations such as Vision Transformers (ViT), attention mechanisms, and multi-scale feature fusion. State-of-the-art methods like TransUNet~\cite{chen2021transunet}, and Swin-UNETR~\cite{hatamizadeh2021swin} have achieved remarkable performance on benchmark datasets for prostate cancer, pancreatic cancer, and other malignancies. However, these models still produce clinically significant failures, such as false positives that could lead to unnecessary biopsies and false negatives that miss critical lesions.

\textbf{Failure detection in medical AI.} Existing failure detection methods rely on output-based uncertainty signals: prediction confidence~\cite{hendrycks2016baseline}, entropy~\cite{hendrycks2016baseline}, or energy scores~\cite{liu2020energy}. While these approaches can flag potential errors, they provide no insight into failure causes and suffer from poor generalization across datasets~\cite{ovadia2019can}. Quality control methods based on image quality assessment~\cite{castro2020causality} or ensemble disagreement~\cite{lakshminarayanan2017simple} similarly lack interpretability. We address this gap by detecting failures leveraging model's internal concepts.

\textbf{Interpretability methods.} Interpretability approaches include gradient-based attribution (GradCAM~\cite{selvaraju2017grad}, IG~\cite{sundararajan2017axiomatic}) and concept extraction via Sparse Autoencoders (SAEs)~\cite{cunningham2023sparse,thasarathan2025universal}. MedSAE applications include cell analysis~\cite{DasMuh_CytoSAE_MICCAI2025}, disease detection~\cite{renzulli2025medsae}, pathology~\cite{le2024learning}, and report generation~\cite{abdulaal2024x}. Concept-based methods (TCAV~\cite{kim2018interpretability}, CBMs~\cite{koh2020concept,yuksekgonul2022post}, medical VLMs~\cite{kim2024transparent,gao2024aligning}) require manual annotation, while unsupervised discovery~\cite{liu2023unsupervised,arefin2024unsupervised} targets natural images. Unlike prior work focused on post-hoc explanation, we leverage unsupervised SAE concepts for actionable failure detection in medical segmentation.

\section{Conclusion, Limitations, and Future Work}
\label{sec:conclusion}

\textbf{Conclusion}. We introduced an interpretable failure detection framework that extracts clinical concepts from segmentation model internals using Sparse Autoencoders. Unlike output-based methods, our approach explains why failures occur by identifying which anatomical concepts drive predictions. Experiments on prostate and pancreatic cancer segmentation show that concept activation patterns effectively distinguish true from false positives, enabling both failure detection and interpretable correction. Critically, our zero-shot transfer results demonstrate that concept-based representations generalize robustly across datasets where confidence-based methods collapse completely.

\textbf{Limitations and future work.} Our method currently captures concept occurrence rather than causal relationships. Understanding how concepts mechanistically interact to produce predictions, rather than merely co-occurring, would enable interventional debugging where clinicians could predict the effect of modifying specific features. Applying causal inference methods such as causal mediation analysis~\cite{meng2022locating,vig2020investigating,imai2010general} or causal abstraction~\cite{geiger2021causal,geiger2025causal,wu2023interpretability} could reveal these mechanistic pathways. Currently, our SAEs are trained and dedicated to each cancer type, requiring retraining for different anatomies. A promising direction is training a universal SAE across multiple cancers (brain, liver, lung, etc) to discover shared anatomical primitives while capturing cancer-specific concepts. This would enable direct comparison of reasoning processes across cancer types and more efficient clinical deployment. Additionally, extending beyond binary failure classification to identify specific failure modes (boundary errors, shape and size estimation failures) would provide more actionable clinical feedback. Finally, one promising direction is to extend our framework to medical segmentation with missing modalities~\cite{karimijafarbigloo2024mmcformer,ma2021smil,ma2022multimodal,azad2022medical,wang2023multi}. By learning modality-specific and shared internal concepts, the model could identify which information is missing and whether the remaining modalities are sufficient for a reliable prediction. These concept-level signals could support failure detection and guide adaptive fusion and modality imputation for robust clinical deployment.

\section*{Acknowledgment} 
\noindent This work is supported by the National Science Foundation under grant numbers CAREER 2340074, SLES 2416937, and III CORE 2412675, the National Institutes of Health under grant number R21CA301093, and the Department of Defense under grant number AFOSR FA9550-23-1-0494. Any opinions, findings, and conclusions or recommendations expressed in this material are those of the authors and do not reflect the views of the supporting entities.


%
%
\bibliographystyle{splncs04}
\bibliography{references}

\newpage

\appendix

\makeappendixtitle{%
Supplementary Material for\\[0.2em]
Medical AI Encodes a ``Feeling of Error'':
Verifying Cancer Segmentation via Internal Concepts
}
\vspace{-20pt}

\setcounter{figure}{8}
\setcounter{table}{5}

\section{Segmentation Failure Detection for Brain Tumor}

\textbf{Data.}  We use the \textit{MSD-Brain}~\cite{antonelli2022medical} dataset, which contains 484 MRI scans with annotations for brain tumors. We employ a split of 310 training, 78 validation, and 96 testing scans. The task involves segmenting three tumor components: the ``complete'' tumor region, the tumor ``core,'' and the ``enhancing'' portion.

\setlength{\intextsep}{2pt}%
\setlength{\columnsep}{10pt}%
\begin{wrapfigure}{r}{0.55\textwidth}
\includegraphics[width=\linewidth]{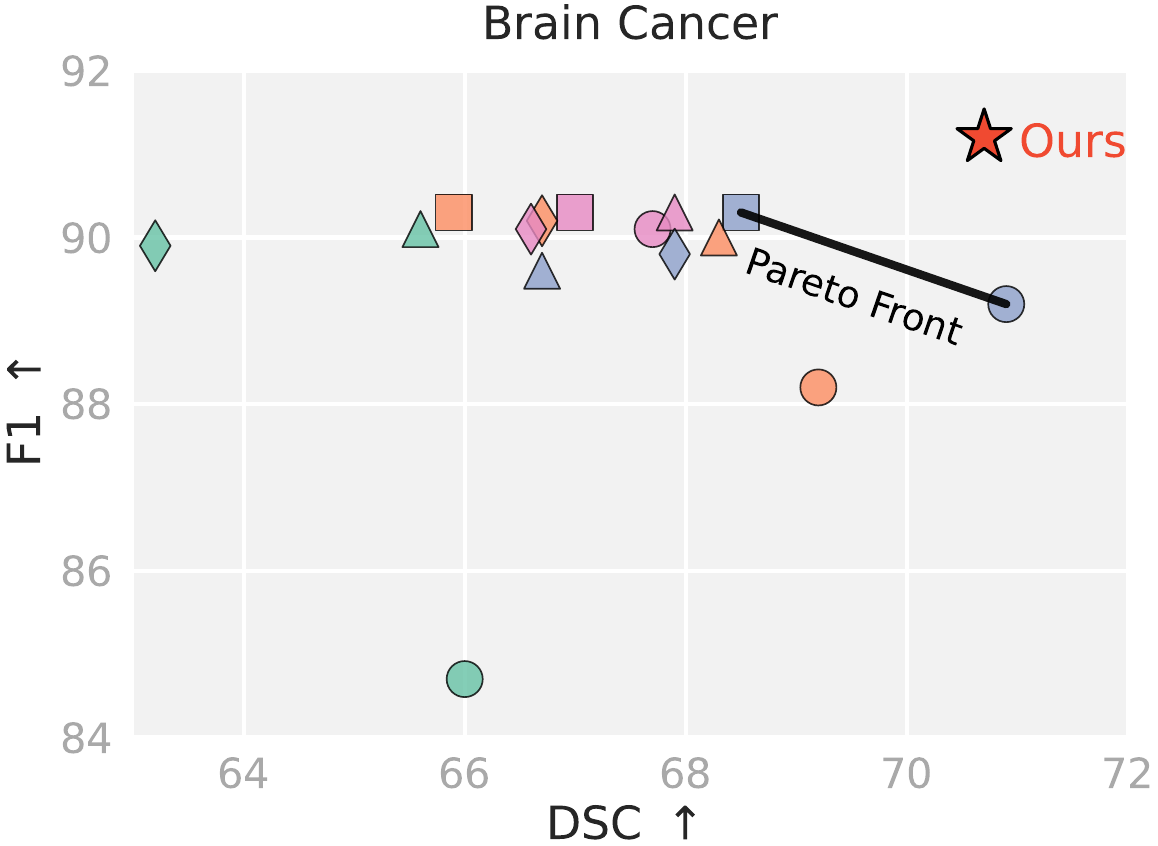}
  \caption{F1 ($\uparrow$) vs. DSC ($\uparrow$) scores on MSD-Brain dataset. Each marker represents a baseline method varying by threshold [0.5, 0.8]. \textit{Our method surpasses the baseline Pareto front, achieving both higher DSC and higher F1, demonstrating superior reliability without sacrificing segmentation quality.}}
  \label{fig:brain}
\end{wrapfigure}

\textbf{Results.} MSD-Brain contains exclusively positive cases, meaning every scan includes brain tumors. This distribution differs markedly from PI-CAI~\cite{saha2024picai} and PanTS~\cite{li2025pants}, which exhibit more natural distributions with a higher proportion of negative (tumor-free) cases relative to positive (tumor-present) cases. Training on this brain tumor dataset reduces both false positives and false negatives, as the balanced representation of tumor versus non-tumor images provides more consistent learning signals. This is reflected in Fig.~\ref{fig:brain}; the F1 scores are substantially higher than the previous two datasets. Notably, our method maintains its effectiveness in this setting, achieving a good balance between segmentation quality and failure detection performance.

\begin{figure}[!th]
  \includegraphics[width=\linewidth]{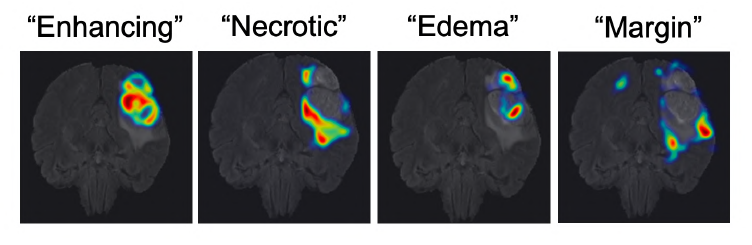}
  \vspace{-20pt}
  \caption{Learned internal concepts and their spatial localization for brain tumor segmentation. Each heatmap shows regions where the model attends to specific tumor characteristics: enhancing regions, necrotic core, edema, and tumor margins.}
  \label{fig:brainconcept}
  \vspace{-10pt}
\end{figure}

\textbf{Visualization}. Fig.~\ref{fig:brainconcept} shows the learned concepts and their spatial localization. The identified concepts are highly relevant to brain tumor characteristics and are accurately localized to the right anatomical structures. Fig.~\ref{fig:bratscorrection} displays examples of detected segmentation failures with the corrected masks generated by our method.  These results align with our findings on prostate and pancreatic tumors, confirming that internal concepts generalize effectively across different cancer types. This consistency demonstrates the robustness of our approach for failure detection in diverse oncological segmentation tasks.

\section{Model Details}

\textbf{Model}. We employ the pretrained MedSAM model~\cite{ma2025rationale}
It typically includes a prompt encoder that requires manual input of a prompt ({\it e.g.}, bounding box). To achieve full automation, we will replace the prompt encoder with an automated “overload encoder”, {\it e.g.}, ResNet~\cite{he2016deep}. This model directly takes images as input and generates surrogate prompts for segmentation, eliminating the need for manual user inputs. Additionally, we modify MedSAM's mask decoder for multiple classes output, enabling simultaneous segmentation of different anatomical structures such as glands, zones, and lesions.

\section{Additional Discussion}

\begin{figure*}[ht]
  \includegraphics[width=\linewidth]{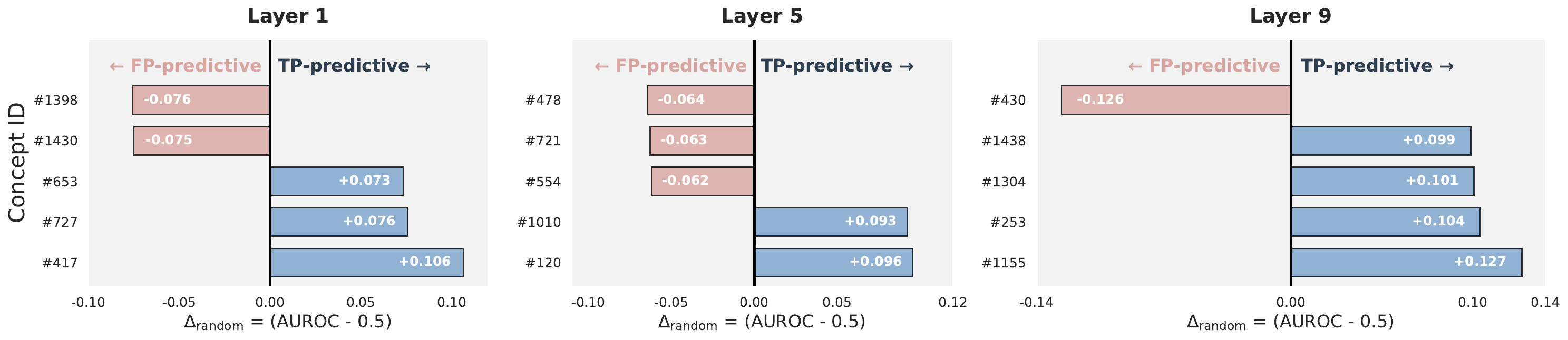}
  \caption{Predictive power of SAE concepts across Layers 1, 5, and 9 for segmentation failures. Each bar shows AUROC improvement over random chance ($\Delta_{\text{random}} = \text{AUROC} - 0.5$) for the top 5 most discriminative concepts per layer. Blue bars indicate concepts predictive of true positives (TP-predictive), while red bars indicate concepts predictive of false positives (FP-predictive). Deep layers exhibit stronger predictive power overall, middle layers show more FP-predictive concepts, and early layers have weaker discriminative ability.}
  \label{fig:localizationNew}
\end{figure*}

\textbf{The predictive power of internal concepts.} In the main text (Fig. 3), we demonstrate the predictive power of concepts in layers 3, 7, and 11 on PI-CAI dataset. Fig.~\ref{fig:localizationNew} extends this analysis to Layers 1, 5, and 9.
To measure each concept's predictive power, we proceed as follows: The SAE decomposes each layer's activations into a fixed set of sparse concepts, each with a non-negative activation value for every image. For each concept, we treat its activation values across all test images as prediction scores and compute the AUROC using ground truth TP/FP labels (where TP indicates correct segmentation and FP indicates segmentation failure). Concepts with $\text{AUROC} > 0.5$ are TP-predictive, meaning higher activation correlates with successful segmentation. Concepts with $\text{AUROC} < 0.5$ are FP-predictive, meaning higher activation correlates with segmentation failures. Fig.~\ref{fig:localizationNew} displays $\Delta_{\text{random}} = |\text{AUROC} - 0.5|$ for the top 5 discriminative concepts per layer, representing predictive power beyond random chance. Our analysis reveals that Layer 9 exhibits stronger discriminative power ($\Delta_{\text{random}} \approx 0.10-0.13$), while Layers 5 and 1 provide complementary information. Notably, Layer 5 is particularly effective at identifying false positive predictions, as evidenced by the prevalence of FP-predictive concepts, whereas deeper layers excel at detecting true positives. This hierarchical pattern suggests that multi-layer concept integration captures diverse failure modes across different levels of feature abstraction.

\begin{wrapfigure}{r}{0.58\textwidth}
    \vspace{-13pt}
    \centering

    \begin{minipage}{\linewidth}
        \centering
        \captionsetup{type=table}
        \caption{Segmentation results of SAE latent amplification.
        ``Raw'' denotes the DSC of the original model mask before intervention.}
        \label{tab:causal}

        \resizebox{\linewidth}{!}{%
        \begin{tabular}{l|c}
        \toprule
            Intervention mode & $\Delta_{\text{DSC}}$ vs.\ raw \\ \midrule
            \textit{reconstruct only}
            (SAE round-trip, no edit)
                & $+0.013$ \\

            \textit{random latent}
            ($\alpha=2$, non-tumor)
                & $+0.035$ \\
            \hline

            \textbf{\textit{amplify \#469}}
            ($\alpha=2$, tumor-related latent)
                & $\mathbf{-0.173}$ \\ \bottomrule
        \end{tabular}%
        }
    \end{minipage}

    \vspace{8pt}

    \begin{minipage}{\linewidth}
        \centering
        \includegraphics[width=\linewidth]{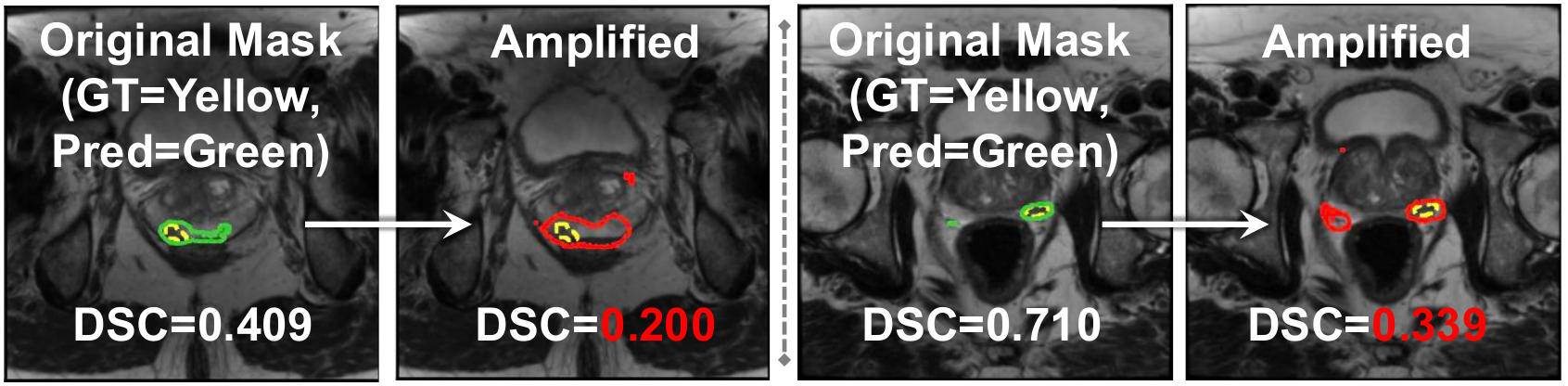}
        \vspace{-15pt}
        \captionsetup{type=figure}
        \caption{DSC after latent amplification. \textit{Amplifying the
        tumor-related latent introduces more false positives and
        substantially reduces DSC.}}
        
        \label{fig:amplify}
    \end{minipage}

\end{wrapfigure}
\textbf{Causal analysis of SAE concepts.} We test causality by \textit{amplifying} the tumor-related concept shown in Fig.~8. For each patch in the image, we encode the Layer 11 activation through the SAE, amplify one latent in the sparse code, decode it, and splice it back into the model. Tab.~\ref{tab:causal} and Fig.~\ref{fig:amplify} show that amplifying tumor-related latent \#469 \textit{sharply reduces DSC}, while the two controlled experiments slightly improve it. This provides direct intervention evidence that \#469 influences the model output, rather than being merely correlated with tumor regions. Interestingly, SAE reconstruction alone slightly improves DSC, suggesting that the SAE may also act as a denoiser through sparse reconstruction.

\begin{wraptable}{r}{0.58\textwidth}
    \vspace{-11pt}
    \centering
    \caption{Comparison with test-time augmentation for failure detection. Compared with TTA, \textit{our method yields better failure detection performance} while also providing concept-level explanations for the detected failures.}
    \label{tab:tta_comparison}
    \vspace{2pt}

    \resizebox{\linewidth}{!}{%
    \begin{tabular}{l|cccc}
    \toprule
        Method & F1$\uparrow$ & Acc$\uparrow$ & DSC & Cost \\
        \midrule
        MedSAM base
            & 0.359
            & 79.55\%
            & 0.513
            & $1\times$ \\

        TTA ($K=6$)
            & 0.592
            & 94.34\%
            & 0.434
            & $6\times$ \\

        \textbf{Ours}
            & \textbf{0.635}
            & \textbf{95.85\%}
            & \textbf{0.501}
            & $\mathbf{1\times}+6$ SAEs \\ \bottomrule
    \end{tabular}%
    }

\end{wraptable}

\textbf{Test-time augmentation.} We added a test-time augmentation (TTA) baseline~\cite{wang2019aleatoric} on the PI-CAI dataset using $K{=}6$ transforms: identity, hflip, vflip, hflip + vflip, and $\pm10\%$ brightness (Tab.~\ref{tab:tta_comparison}). 
We use slice disagreement, defined as $1-$mean pairwise DSC across TTA predictions, as the failure score and sweep the threshold for best F1. 
Our SAE-based detector outperforms TTA in F1, accuracy, and DSC after failure correction, while requiring one model forward pass plus SAE encoding rather than six full model passes. 
We also found that TTA saturates at $K{=}6$: reducing to $K{=}4$ gives nearly identical F1 ($0.591$).

\begin{figure}[!ht]
\centering
  \includegraphics[width=\linewidth]{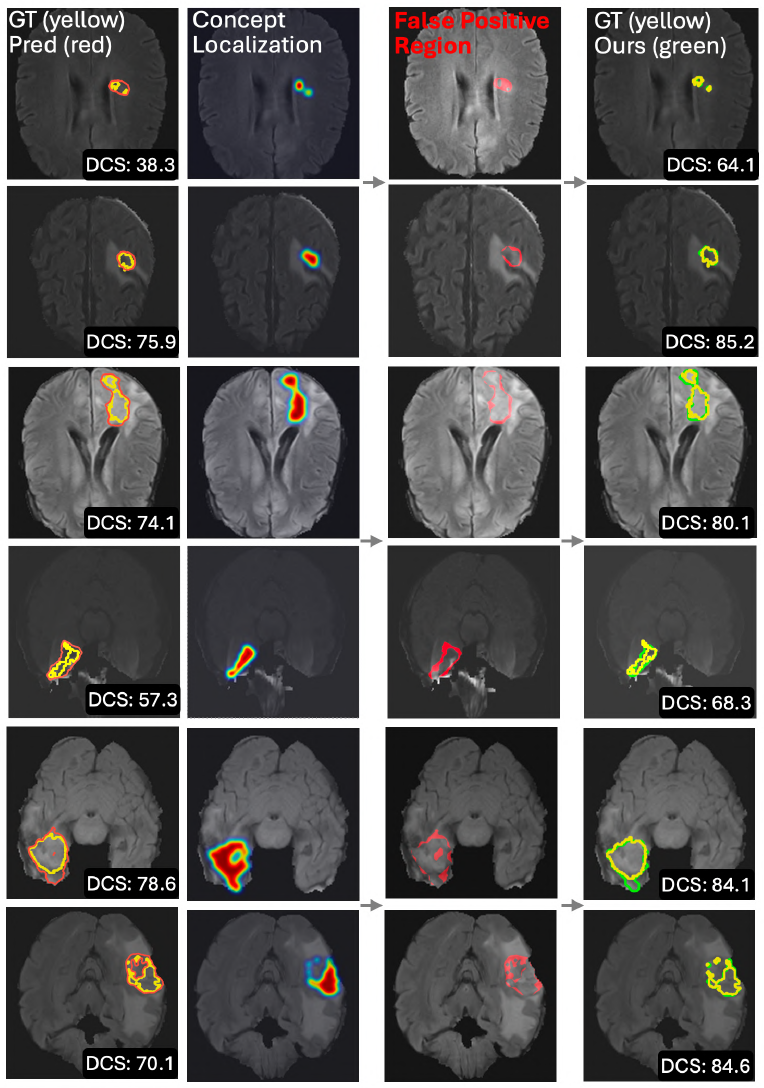}
  \caption{Visualization of FP correction on MSD-Brain dataset. \textit{Left:} Ground truth (yellow), initial prediction (red), and internal concept localization highlighting regions contributing to the prediction. \textit{Middle}: Detected false positive regions in red.  \textit{Right}: Final prediction mask (green) after FP removal.}
  \label{fig:bratscorrection}
\end{figure}

\end{document}